\documentclass[letterpaper, 10 pt, conference]{ieeeconf}  

\IEEEoverridecommandlockouts                              

\usepackage{times}
\usepackage{soul}
\usepackage{url}
\usepackage[hidelinks]{hyperref}
\usepackage[utf8]{inputenc}
\usepackage{caption}
\usepackage{graphicx}
\usepackage{amsthm}
\usepackage{amsmath}
\usepackage{booktabs}
\usepackage{algorithm}
\usepackage{algorithmic}
\usepackage[switch]{lineno}
\usepackage{amsmath}
\usepackage{amssymb}

\usepackage{xcolor}
\usepackage{comment}
\usepackage{wrapfig}
\usepackage{relsize}
\usepackage{multirow}

\DeclareMathOperator*{\argmin}{arg\,min}

\title{\LARGE \bf
BucketKD: A Safety-Aware Bucket-Based Knowledge Distillation Framework for End-to-End Motion Planning}

\author{
    Md Nahidul Islam$^{2}$, Mohd Hasan Ali$^{1}$, Dipankar Dasgupta$^{2}$, and Myounggyu Won$^{2}$ \\
    \thanks{$^{1}$Mohd Hasan Ali is with the Department of Electrical and Computer Engineering, University of Memphis, Memphis, TN, United States
        {\tt\small mhali@memphis.edu}}%
    \thanks{$^{2}$Md Nahidul Islam, Dipankar Dasgupta, and Myounggyu Won are with the Department of Computer Science, University of Memphis, Memphis, TN, United States
        {\tt\small \{mislam19, ddasgupt, mwon\}@memphis.edu}}%
}

\begin{document}

\maketitle

\begin{abstract}

End-to-end motion planning has emerged as a promising paradigm in autonomous driving, directly mapping raw sensor data to control commands via deep neural networks. Despite its advantages, its large model size hinders deployment in resource-constrained platforms. In this paper, we present BucketKD, a bucket-based knowledge distillation framework that yields compact and safety-aware end-to-end planners. Compared to the state-of-the-art approach, which relies on simplified planning state representations, BucketKD discretizes critical environmental variables into adaptive buckets that capture richer scene semantics while preserving efficiency. In addition, we design a safety-aware waypoint attention mechanism that evaluates each waypoint’s risk level by accounting for both obstacle proximity and relative motion through a time-to-collision (TTC) formulation widely used in transportation research. This enables the student model to better retain safety-critical behaviors during distillation. Extensive experiments in CARLA using the Bench2Drive dataset show that BucketKD significantly outperforms the state-of-the-art in both planning accuracy and safety while maintaining strong compression ratios.

\end{abstract}

\section{Introduction}
\label{sec:introduction}

End-to-end motion planning has emerged as a promising paradigm in autonomous driving. In this approach, deep neural networks learn an end-to-end mapping from raw sensor data to the vehicle’s trajectory or control actions, removing the need for explicit perception, planning, and control submodules~\cite{zeng2019end,feng2024road,zhang2025bridging,song2025don}. End-to-end motion planning offers reduced latency, greater efficiency, and stronger generalization than traditional modular pipelines by integrating perception, prediction, and planning into a single data-driven model~\cite{teng2023motion,zheng2025world4drive}. However, a key challenge of this approach is its large model size, which typically demands high-performance onboard hardware, making deployment difficult in resource-constrained environments such as autonomous delivery robots or vehicles with limited computational capacity~\cite{feng2024road}. 

To mitigate the model size limitation, a state-of-the-art compression framework PlanKD leverages knowledge distillation to achieve compact yet high-performing end-to-end motion planners~\cite{feng2024road}. Inspired by the bottleneck principle~\cite{alemi2016deep}, PlanKD learns compact representations that capture planning-critical features with minimal dimensionality while maintaining sufficient information for robust decision-making. While PlanKD achieves substantial model compression without notable performance degradation, the oversimplification of its planning state representation constrains its capability to handle complex and dynamic driving scenarios. Furthermore, its safety model’s dependence on distance-based metrics alone limits its ability to assess the full contextual risk of the environment, thereby increasing the likelihood of suboptimal or unsafe maneuvers.

In this paper, we present \textbf{BucketKD}, a \textbf{Bucket}-based \textbf{K}nowledge \textbf{D}istillation framework for end-to-end motion planning in autonomous driving. Given the limited generalization of binary state representations in existing knowledge distillation frameworks, especially under complex and dynamically evolving traffic conditions, we introduce a bucket-based method that discretizes each critical state variable into a set of nonuniform buckets to capture finer-grained behavioral variations. This design allows the planning states to capture the complexity of the driving environment more effectively, while keeping the total number of states manageable. By maintaining a small number of buckets per state, our method preserves compression efficiency without sacrificing planning performance. Additionally, in the knowledge distillation process, we incorporate an attention mechanism to compute attention weights between each waypoint and its corresponding context in the bird's-eye-view (BEV) representation of the driving scene. These attention weights are determined not only based on the distance to nearby moving obstacles, but also by considering their relative motion—specifically, whether the obstacles are approaching and how quickly they are closing in. This enables the model to more accurately identify safety-critical waypoints, thereby improving the effectiveness of knowledge distillation and enhancing overall safety in the distilled model. We conduct extensive experiments in CARLA on the standard Bench2Drive dataset and observe that BucketKD consistently surpasses the state-of-the-art method in both safety and overall planning performance.

Our key contributions are summarized as follows.

\begin{itemize}
	\item We develop a novel bucket-based formulation of planning states that bridges the gap between model compactness and situational awareness, yielding improved performance in complex and dynamic driving conditions.
	\item We design an adaptive safety-aware knowledge distillation method that prioritizes safety-critical waypoints by analyzing the relative motion of surrounding obstacles, improving planning accuracy in high-risk scenarios.
	\item We conduct extensive evaluations on the standardized Bench2Drive benchmark in CARLA, comparing against state-of-the-art teacher and student models to validate the effectiveness and robustness of the proposed approach across diverse and challenging driving scenarios.
\end{itemize}

This paper is organized as follows. Section~\ref{sec:related_work} reviews the literature on end-to-end motion planning and knowledge distillation methods for end-to-end motion planning. Section~\ref{sec:preliminaries} presents the preliminaries required for our work. Section~\ref{sec:our_approach} describes the proposed BucketKD framework, including the bucket-based planning-relevant feature distillation (BPFD) and safety-aware waypoint-attentive distillation (SWD) modules. Section~\ref{sec:results} presents the simulation results and corresponding analysis. Finally, Section~\ref{sec:conclusion} concludes the paper.

\section{Related Work}
\label{sec:related_work}

End-to-end motion planning is an emerging paradigm in autonomous driving that maps raw sensor data directly to final trajectories or control commands~\cite{zeng2019end,feng2024road,zhang2025bridging}.  This learning-based paradigm reduces reliance on extensive hand-crafted rules and alleviates the compounding of errors that typically occurs across multiple stages in traditional modular architectures~\cite{prakash2021multi,xu2017end}. 

Numerous studies have been conducted on end-to-end motion planning. Zhang \emph{et al.}~\cite{zhang2021end} developed a reinforcement learning (RL) expert, Roach, which maps bird’s-eye view (BEV) images to continuous low-level actions, providing rich supervision signals for training imitation learning agents. Wang \emph{et al.}~\cite{wang2021learning} proposed an interpretable end-to-end motion planning framework that leverages visual inputs for autonomous driving, incorporating an optical flow distillation mechanism to improve performance without compromising real-time capability. Wu \emph{et al.}~\cite{wu2022trajectory} proposed an integrated learning framework that simultaneously optimizes trajectory planning and direct control modules within a unified end-to-end autonomous driving system. Shao \emph{et al.}~\cite{shao2023safety} enhanced the safety and interpretability of end-to-end motion planning by introducing a unified one-stage architecture that efficiently fuses information from multiple sensor modalities. Building on this line of research, Shao \emph{et al.}~\cite{shao2024lmdrive} leveraged large language models to strengthen reasoning ability, interpretability, and overall system performance for autonomous vehicles in closed-loop environments. Recognizing the limitations of existing evaluation methodologies for end-to-end autonomous driving, Jia \emph{et al.}~\cite{jia2024bench2drive} proposed Bench2Drive, a comprehensive benchmark designed to enable realistic and fair closed-loop assessments of such systems. Similarly, Li \emph{et al.}~\cite{li2024ego} highlighted that trajectory outputs are strongly affected by ego-vehicle states (e.g., velocity, acceleration, yaw angle) and advocated for richer and more diverse evaluation metrics to prevent models from overfitting to specific objectives. In addition, Yu \emph{et al.}~\cite{yu2025end} developed a framework that leverages V2X communication to fuse infrastructure and ego-vehicle sensor data, thereby improving planning accuracy and overall driving performance.

Despite their impressive success driven by deep neural networks, end-to-end motion planning models often suffer from excessive model complexity and high computational requirements, making them impractical for real-time deployment in resource-limited autonomous systems~\cite{feng2024road}. To mitigate the challenges posed by large model sizes, Feng~\emph{et al.}~\cite{feng2024road} introduced a compression framework leveraging knowledge distillation~\cite{hong2022cross} guided by the information bottleneck principle~\cite{alemi2016deep}, enabling compact end-to-end motion planning models that retain high performance. More specifically, their approach utilizes knowledge distillation to isolate planning-critical features, retaining the minimal yet sufficient representations required to support accurate and reliable motion planning. While the approach yields efficient compression, its dependence on coarse binary planning states hampers situational awareness and degrades performance under complex and dynamic traffic conditions. Moreover, the method evaluates safety-critical waypoints exclusively through distance-based metrics, disregarding the relative motion dynamics between the ego-agent and surrounding obstacles, which leads to suboptimal and less reliable safety evaluation. This paper tackles these limitations by developing a bucket-based knowledge distillation framework that enhances state representation and motion-aware safety reasoning to achieve robust end-to-end motion planning.

\section{Preliminaries}
\label{sec:preliminaries}

An end-to-end motion planning framework bypasses traditional modular pipelines by processing raw sensor data such as LiDAR, radar, and camera inputs through deep neural networks to directly generate the vehicle’s planned trajectory or low-level control actions. Formally, a motion planning model takes as input the state 
$I = \{o, m, c\}$, where $o$ represents the raw sensory observations 
(\emph{e.g.}, camera images), $m$ denotes the ego-vehicle state 
(\emph{e.g.}, velocity, acceleration, yaw angle), and $c$ specifies the 
high-level driving command (\emph{e.g.}, turn left, turn right, go straight, follow). The motion planning model produces either a continuous trajectory 
that defines the vehicle's intended motion or a set of low-level 
control actions (\emph{e.g.}, steering angle, throttle, braking) obtained 
through a trajectory-tracking controller such as a PID regulator 
using the predicted trajectory as reference. In this study, we employ imitation learning~\cite{codevilla2018end} 
to train our motion planning framework. This data-driven approach, 
commonly used in end-to-end autonomous driving, enables the model 
to learn by replicating expert demonstrations and to generate 
trajectories that align with expert driving behaviors. 
The corresponding learning objective is formulated as follows.

\begin{equation}
	\argmin_{\theta}\mathbb{E}_{(I, \mathcal{T}^*)\sim\mathcal{D}}[\mathcal{L}(\mathcal{F}_{\theta}(I), \mathcal{T}^*)],
\end{equation}

\noindent where $\mathcal{D} = (I, \mathcal{T}^*)$ represents the training dataset, 
with each entry containing the input state $I$ and its corresponding 
expert trajectory $\mathcal{T}^*$. The expert trajectory 
$\mathcal{T}^*$ consists of $T$ waypoints, \emph{i.e.}, 
$\mathcal{T}^* = \{w_i^*\}_{i=1}^{T}$, where $w_i^*$ denotes the 
$i$-th waypoint. The motion planner, parameterized by $\theta$, is 
expressed as $\mathcal{F}_{\theta}$. The loss function $\mathcal{L}$ 
quantifies the deviation between the predicted trajectory and the 
expert trajectory using the $L_1$ norm, corresponding to the mean 
absolute error.

The state-of-the-art knowledge distillation framework for motion 
planning leverages distillation to mitigate the excessive parameter 
growth inherent in large end-to-end networks. Specifically, the 
teacher model is represented as $\mathcal{F}^{T}_{\theta}$ and the 
student model as $\mathcal{F}^{S}_{\phi}$, with $\theta$ and $\phi$ 
denoting their respective parameter sets. The distillation process 
aims to transfer critical task-relevant knowledge from the teacher 
to the student, enabling the student model to achieve comparable 
performance while maintaining a significantly reduced computational 
and memory footprint.

\section{Design of BucketKD}
\label{sec:our_approach}

\subsection{Overview}
\label{sec:overview}

\begin{figure*}
	\centering
	\includegraphics[width=0.95\textwidth]{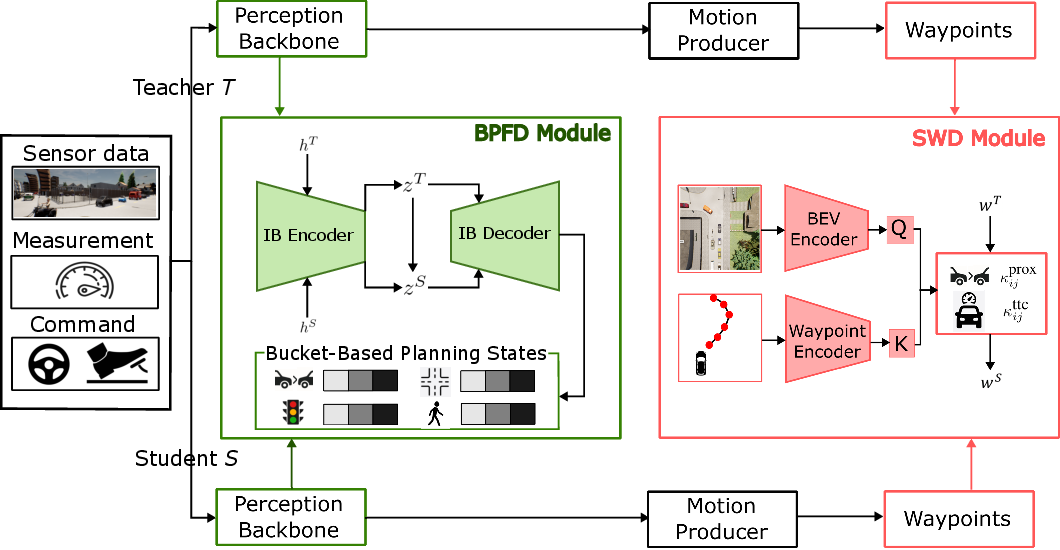}
	\caption{The BucketKD architecture. It consists of two main components: the bucket-based planning-relevant feature distillation (BPFD) module and safety-aware waypoint-attentive distillation (SWD) module. $h^T$ and $h^S$ denote the intermediate features of the teacher and student models, respectively ($z^T$ and $z^S$ are the corresponding planning-relevant features). $w^T$ and $w^S$ are the waypoints for the teacher and student models, respectively.}
	\label{fig:overview}
\end{figure*}

A standard end-to-end motion planning network integrates two principal subsystems: the perception backbone and the motion producer~\cite{tampuu2020survey}. The perception backbone interprets raw sensory inputs to extract and encode salient environmental features, while the motion producer leverages these encoded representations to compute the planned trajectory~\cite{wu2022trajectory}.

Following this architectural structure, BucketKD introduces two complementary knowledge distillation modules, each tailored to one of the key components of an end-to-end motion planner, as illustrated in Fig.~\ref{fig:overview}. The first module, the \textbf{B}ucket-based \textbf{P}lanning-relevant \textbf{F}eature \textbf{D}istillation (\textbf{BPFD}), targets the perception backbone to extract high-level planning cues from its intermediate feature representations. Unlike the state-of-the-art distillation scheme~\cite{feng2024road} that relies on coarse or binary state abstractions, BPFD implements bucket-based planning state representation that discretizes key environmental and dynamic variables into adaptive buckets. This formulation preserves fine-grained semantic and spatial information, enabling the student network to capture subtle variations in traffic context—such as relative positioning, obstacle motion, and local geometry—that are often lost in simplified encodings. By doing so, BPFD significantly strengthens the fidelity of knowledge transfer between teacher and student networks and enhances robustness in complex, dynamically changing driving environments where nuanced scene understanding is essential for safe and reliable trajectory generation.

The second module, the \textbf{S}afety-aware \textbf{W}aypoint-attentive \textbf{D}istillation (\textbf{SWD}) module, operates on the motion producer to enhance the transfer of safety-critical knowledge. This module dynamically assigns adaptive importance weights to each waypoint according to its estimated safety criticality, ensuring that the student model learns to prioritize decision points that are more relevant to collision avoidance and safe maneuvering. To achieve this, motivated by the significant transporation safety literature~\cite{minderhoud2001extended,vogel2003comparison,zhu2020safe}, we introduce a  Time-to-Collision (TTC)-based waypoint weighting mechanism that extends beyond the distance-only criterion used in prior work~\cite{feng2024road}. Instead of relying solely on the proximity between the ego vehicle and surrounding objects, the proposed scheme explicitly models their relative motion, incorporating both directionality and relative velocity to quantify potential risk. By considering whether nearby obstacles are approaching and how quickly they do so, the SWD module produces a more realistic and context-aware estimation of safety-critical waypoints. This, in turn, improves the fidelity of the distillation process, enabling the student model to better capture fine-grained temporal safety cues and perform more reliably in high-risk, safety-sensitive driving situations.

\subsection{Bucket-based Planning-relevant Feature Distillation (BPFD)}

The objective of knowledge distillation in end-to-end motion planning is to selectively capture and transfer information that is directly relevant to planning decisions, while suppressing redundant or non-essential scene details~\cite{cho2019efficacy,park2019relational}. To achieve this, the process leverages the information bottleneck (IB) principle~\cite{feng2024road}, which constrains the distilled representation to retain only the minimal yet sufficient set of planning-critical features. This principle effectively filters out visual or contextual noise such as background structures or distant non-interactive agents that do not influence driving behavior, thereby enhancing the efficiency and interpretability of the learned model. More specifically, the information bottleneck (IB) principle 
aims to learn a compact latent representation $Z$ of the input $X$ 
that retains only the information necessary for accurately predicting 
the target variable $Y$~\cite{tishby2000information}. In the context of 
end-to-end motion planning, the goal is to learn a 
representation that encapsulates only the driving-relevant cues from 
sensory inputs while discarding extraneous scene details. Accordingly, 
the problem can be formulated from the IB perspective as 
follows~\cite{feng2024road}.

\begin{equation}
	\label{eq_ib}
	J_{IB} = \max_{Z} \sum_{i=1}^{M} I(Z, Y^i) - \beta I(Z, X),
\end{equation}

\noindent \noindent where $M$ denotes the number of planning states, each 
representing a compact set of semantic labels that encapsulate both 
the driving-relevant context and the ego vehicle's short-term control 
intent. $I(\cdot, \cdot)$ denotes the mutual information operator; 
$Y^i$ is the ground-truth label corresponding to the $i$-th planning 
state; $\beta$ is the Lagrange multiplier that governs the trade-off 
between information compression and task relevance; and $X$ represents 
the intermediate feature map extracted from the perception backbone.

Eq.~(\ref{eq_ib}) highlights that the overall effectiveness of the 
information bottleneck framework is largely determined by the design 
of the planning states. Although a simple binary representation~\cite{feng2024road} 
offers ease of implementation and interpretability, it oversimplifies 
the rich spatial and contextual knowledge encoded in the teacher model. 
Such a coarse abstraction may fail to convey fine-grained environmental 
cues that are critical for safe and efficient planning. For instance, 
in real driving scenarios, the decision regarding ``nearby vehicles'' 
cannot be reduced merely to their presence or absence—optimal maneuvering 
often depends on more nuanced factors such as the number of surrounding 
vehicles, their spatial density, and relative configuration. Consequently, 
a more expressive representation of planning states is essential to 
fully exploit the teacher's informative latent features and enhance the 
student's situational awareness.

To bridge this representational gap, BucketKD incorporates a bucket-based 
planning state formulation that captures richer contextual semantics 
than binary encodings. Unlike the coarse two-level representation that 
merely indicates the presence or absence of specific conditions, the 
bucket-based approach discretizes continuous variables into multiple 
adaptive intervals, or ``buckets.'' This formulation allows the student model to 
inherit a more nuanced understanding of spatial and temporal 
relationships from the teacher, thus enhancing its decision-making 
fidelity in diverse driving contexts. For example, we categorize 
the planning states into environmental planning states (E) and 
action planning states (A), defined as follows, which can be easily modified depending on the use of the approach.

\begin{itemize}
	\item Nearby vehicles (E): None / Single / Few (2–3) / Many (4+)
	\item Pedestrians (E): None / Crosswalk-involved / Roadside
	\item Traffic lights (E): Absent / Red / Yellow / Green
	\item Junction (E): No / Approaching / In-junction
	\item Action States (A): brake (none/mild/strong), throttle (none/mild/strong), steer (straight/mild/strong), lane-change intent (none/left/right).
\end{itemize}

After the teacher's information bottleneck (IB) module produces a 
compact latent representation $Z^T$ that encapsulates the 
planning-relevant information, the student network is trained to 
reconstruct this representation through feature-level alignment. 
To achieve this, the student learns to minimize the discrepancy 
between its encoded features $Z^S$ and the teacher's distilled 
output $Z^T$ using an $L_1$ loss function, which enforces consistency 
in the learned latent space and promotes faithful knowledge transfer. 
The loss function is defined as follows:

\begin{equation}
	\mathcal{L}_z = \frac{1}{N} \sum_{i=1}^{N} \left| z_i^T - z_i^S \right|,
\end{equation}

\noindent where $N$ denotes the number of samples.

\subsection{Safety-Aware Waypoint-Attentive Distillation (SWD)}

In motion planning, the importance of each waypoint varies with the driving context. To model this effect, we introduce attention weights that capture the relationship between the BEV scene image $B \in \mathbb{R}^{C \times H \times W}$  and each waypoint $w_i$ in a trajectory $\mathcal{T} = \{ w_i \in \mathbb{R}^2 \}_{i=1}^T$. The attention weight is calculated as follows. 

\begin{equation}
	Q = f_{bev}(\tilde{B}),
	K = f_{w}(\mathcal{T}),
	A = \text{softmax}\!\left(\frac{QK}{\sqrt{d_k}}\right),
\end{equation}

\noindent where $a_i \in A = \{a_i\}_{i=1}^T$ is the attention weight of waypoint $w_i$, and $d_k$ is the dimension of $K$. We adopt the BEV encoder $f_{bev}$ and the waypoint encoder $f_w$ from~\cite{feng2024road}. 

BucketKD prioritizes safety-critical waypoints by imposing higher weights on them, thereby enhancing the retention of safety-aware information in the distillation stage. We note that relying solely on the proximity of nearby moving obstacles, as done in the state-of-the-art method~\cite{feng2024road}, is insufficient for accurately assessing waypoint safety. More specifically, while closeness to surrounding objects (often neighboring vehicles) is important, a reliable safety evaluation also requires considering whether those objects are moving toward or away from the ego vehicle and at what speed. To assess the safety level more accurately in distilling the waypoints, we propose the following safety kernal function $\psi_i$.

\begin{equation}
	\psi_i = \mathlarger\sum_j \kappa_{ij}^{\mbox{prox}} \cdot  \mathlarger\sum_j \kappa_{ij}^{\mbox{ttc}},
\end{equation}

\begin{equation}
	\kappa_{ij}^{\mbox{prox}} = e^{-\frac{1}{2\sigma^{2}} \| p_i - p_j \|^{2}},
\end{equation}

\begin{equation}
	\kappa_{ij}^{\mbox{ttc}} = e^{ -\left( \frac{\widehat{\mathrm{TTC}}_{ij}}{\tau} \right)^{\beta} },
\end{equation}

\noindent where $p_i$ and $p_j$ denote the the 2D positions of waypoint $i$ and obstacle $j$, respectively, and $v_i$ and $v_j$ denote their corresponding velocities; $\tau$ is a scaling parameter; and $\beta$ controls sharpness of decay. $\widehat{\mathrm{TTC}}_{ij}$ is the estimated time-to-collision between waypoint $i$ and obstacle $j$, which is defined as follows. 

\begin{equation}
	\widehat{\mathrm{TTC}}_{ij} =
	\begin{cases}
		\dfrac{\| p_i - p_j \|}{s_{ij} + \varepsilon}, & s_{ij} > 0, \\[1.0em]
		\infty, & s_{ij} \le 0,
	\end{cases}
\end{equation}

\noindent where $s_{ij} = -\dfrac{(p_j - p_i)^\top (v_j - v_{i})}{\| v_j - v_{i} \| + \varepsilon}$, and $\epsilon$ is a small constant to avoid division by zero. $s_{ij} > 0$ indicates that the obstacle is approaching the ego vehicle, and a larger $s_{ij}$ corresponds to a higher closing speed and thus greater safety importance. On the other hand, if $s_{ij} \le 0$ (not approaching), $\widehat{\mathrm{TTC}}_{ij}$ becomes large $\rightarrow$ kernel $\approx 0$.

Consequently, we design a pairwise ranking loss that encourages waypoints with higher $\psi_i$ values to receive proportionally larger attention weights.

\begin{equation}
	\mathcal{L}_{rank} = \sum_{i=1}^{T} \sum_{j=1}^{T} \max\!\left(0,\,-r_{ij}(a_i - a_j)\right),
\end{equation}

\noindent where the comparison indicator $r_{ij}$ is defined as $r_{ij} = 1$ if $\psi_i > \psi_j$, and $r_{ij} = -1$ otherwise. The variable $a_i$ denotes the attention weight assigned to each waypoint $w_i$. By minimizing $\mathcal{L}_{rank}$, the model learns to assign attention weights that accurately reflect the relative importance of waypoints while incorporating safety considerations.

After computing the attention weights, we incorporate them into the 
imitation learning objective to emphasize safety-critical waypoints 
during distillation. The resulting safety-aware waypoint-attentive 
loss is formulated as  

\begin{equation}
	\mathcal{L}_{w} = \sum_{i=1}^{T} a_i \lvert w_i^S - w_i^T \rvert,
\end{equation}

\noindent where $\mathcal{L}_{w}$ represents the weighted imitation loss, 
and $w_i^S$ and $w_i^T$ denote the waypoints predicted by the student 
and teacher planners, respectively. This formulation ensures that 
waypoints with higher safety importance—reflected through larger $a_i$ 
values—contribute more to the training objective. Following the 
strategy in~\cite{feng2024road}, we additionally introduce an entropy 
regularization term to prevent the model from overemphasizing a few 
waypoints and to promote a more balanced attention distribution:  

\begin{equation}
	\mathcal{L}_{e} = \sum_{i=1}^{T} a_i \log(a_i).
\end{equation}

\noindent This regularization encourages diversity in the attention 
weights, ensuring that the student model maintains adequate focus 
across all relevant waypoints while still prioritizing safety-critical ones.

\subsection{Optimization}
\label{sec:optimization}

The proposed framework is optimized through a unified end-to-end 
objective that jointly balances all learning components. The overall 
loss function is expressed as  

\begin{equation}
	\mathcal{L} = \mathcal{L}_{w} + \mathcal{L}_{w^*} - \mathcal{L}_{IB} 
	+ \alpha_z \mathcal{L}_z + \alpha_r \mathcal{L}_{rank} + \alpha_e \mathcal{L}_e,
\end{equation}

\noindent where $\alpha_z$, $\alpha_r$, and $\alpha_e$ are weighting 
coefficients that regulate the influence of the corresponding terms. 
Here, $\mathcal{L}_{w^*}$ denotes the $L_1$-based imitation loss between 
the predicted and expert trajectories, which provides direct supervision 
and is modulated by the safety-aware attention mechanism—similar to 
$\mathcal{L}_w$. The information bottleneck term $\mathcal{L}_{IB}$ 
corresponds to the lower bound of the IB objective and is maximized to 
encourage the encoder–decoder pair to retain only the most task-relevant 
features. Minimizing the composite loss $\mathcal{L}$ allows the student 
planner to simultaneously inherit high-level perceptual understanding 
and motion generation capabilities from the teacher network, resulting 
in a compact yet reliable end-to-end planner capable of safe and 
context-aware trajectory generation.

\section{Experiments}
\label{sec:results}

\subsection{Experimental Setup}
\label{sec:experimental_setup}

\textbf{Simulation environment.} The BucketKD framework is implemented and tested in CARLA (version 0.9.10.1)~\cite{carla2017}, a high-fidelity simulator extensively used for autonomous driving research due to its realistic perception, traffic, and environment modeling. We perform all training and evaluation on our university HPC cluster. Each experiment is executed on a single NVIDIA A100 80GB PCIe GPU (CUDA 12.3), with 16 CPU cores and 160 GB of allocated system memory. 

\begin{table*}[t]
	\centering
	\small
	\setlength{\tabcolsep}{6pt}
	\begin{tabular}{l c c c c c c c c}
		\toprule
		\multirow{2}{*}{Backbone}
		& \multirow{2}{*}{Param Count}
		& \multirow{2}{*}{\shortstack{Inference\\Time (ms)}}
		& \multirow{2}{*}{\shortstack{With\\BucketKD}}
		& \multicolumn{5}{c}{Metrics} \\
		\cmidrule(lr){5-9}
		& & & 
		& \shortstack{Driving\\Score ($\uparrow$)}
		& \shortstack{Route\\Completion ($\uparrow$)}
		& \shortstack{Infraction\\Score ($\uparrow$)}
		& \shortstack{Collision\\Rate ($\downarrow$)}
		& \shortstack{Infraction\\Rate ($\downarrow$)} \\
		\midrule
		
		\multirow{7}{*}{TCP}
		
		& 26.6M & 113.65 & --
		& 54.14 & 80.92 & 0.778 & 0.594 & 1.548 \\
		\cmidrule(lr){2-9}
		
		& 16.6M & 78.49 & no
		& 29.61 & 59.57 & 0.611 & 0.829 & \textbf{2.172} \\
		& 16.6M & 89.11 & yes
		& \textbf{37.26} & \textbf{68.28} & \textbf{0.643} & \textbf{0.678} & 2.328 \\
		\cmidrule(lr){2-9}
		
		& 10.3M & 37.38 & no
		& 14.05 & 38.64 & 0.429 & 1.954 & 3.537 \\
		& 10.3M & 51.56 & yes
		& \textbf{25.78} & \textbf{49.89} & \textbf{0.596} & \textbf{1.487} & \textbf{2.941} \\
		\cmidrule(lr){2-9}
		
		& 5.4M & 21.79 & no
		& 8.46 & 18.53 & 0.335 & 4.961 & 6.512 \\
		& 5.4M & 29.92 & yes
		& \textbf{16.53} & \textbf{31.92} & \textbf{0.478} & \textbf{3.686} & \textbf{4.879} \\
		
		\bottomrule
	\end{tabular}
	\caption{Performance comparison of TCP models with and without BucketKD across different model sizes on the Bench2Drive dataset.}
	\label{tab:tcp}
\end{table*}

\begin{table*}[t]
	\centering
	\small
	\setlength{\tabcolsep}{6pt}
	\begin{tabular}{l c c c c c c c c}
		\toprule
		\multirow{2}{*}{Backbone}
		& \multirow{2}{*}{\shortstack{Param\\Count}}
		& \multirow{2}{*}{\shortstack{Inference\\Time (ms)}}
		& \multirow{2}{*}{KD Model}
		& \multicolumn{5}{c}{Metrics} \\
		\cmidrule(lr){5-9}
		& & &
		& \shortstack{Driving\\Score ($\uparrow$)}
		& \shortstack{Route\\Completion ($\uparrow$)}
		& \shortstack{Infraction\\Score ($\uparrow$)}
		& \shortstack{Collision\\Rate ($\downarrow$)}
		& \shortstack{Infraction\\Rate ($\downarrow$)} \\
		\midrule
		
		\multirow{8}{*}{TCP}
		
		& 26.6M & 113.65 & BucketKD Teacher
		& \textbf{54.14} & \textbf{80.92} & \textbf{0.778} & 0.594 & \textbf{1.548} \\
		& 26.6M & 104.29 & SOTA Teacher
		& 43.89 & 71.95 & 0.689 & \textbf{0.533} & 1.997 \\
		\cmidrule(lr){2-9}
		
		& 16.6M & 89.11 & BucketKD Student
		& \textbf{37.26} & \textbf{68.28} & \textbf{0.643} & \textbf{0.678} & \textbf{2.328} \\
		& 16.6M & 84.32 & SOTA Student
		& 30.57 & 55.13 & 0.514 & 0.748 & 2.785 \\
		\cmidrule(lr){2-9}
		
		& 10.3M & 51.56 & BucketKD Student
		& \textbf{25.78} & \textbf{49.89} & \textbf{0.596} & \textbf{1.487} & \textbf{2.941} \\
		& 10.3M & 44.71 & SOTA Student
		& 19.31 & 43.22 & 0.527 & 2.529 & 3.713 \\
		\cmidrule(lr){2-9}
		
		& 5.4M & 29.92 & BucketKD Student
		& \textbf{16.53} & \textbf{31.92} & 0.478 & \textbf{3.686} & \textbf{4.879} \\
		& 5.4M & 24.48 & SOTA Student
		& 10.28 & 19.43 & \textbf{0.491} & 4.314 & 6.496 \\
		
		\bottomrule
	\end{tabular}
	\caption{Comparison of BucketKD and SOTA (Teacher and Student models) on the TCP backbone across different model sizes using the Bench2Drive dataset.}
	\label{tab:sota}
\end{table*}

\textbf{Dataset.} we evaluate our approach on the official Bench2Drive benchmark~\cite{bench2drive2024}. It provides a long-horizon, closed-loop, multi-agent evaluation setting which correlates directly to the deployment conditions of motion planning systems. The dataset is generated using CARLA version 0.9.15 and reinforcement learning based expert model Think2Drive~\cite{leonardisthink2drive2025} in 10 Hz frame rate. The training dataset comprises 2 million fully annotated frames from 13638 clips distributed uniformly over 44 interactive scenarios under 23 weather conditions and 12 towns. The evaluation dataset consists of 220 short routes which are around 150 meters long with a single specific scenario. We use Roach Agent~\cite{zhang2021end} to evaluate the models following standard Bench2Drive evaluation protocol.

\textbf{Evaluation metrics.} We evaluate our approach using three widely adopted metrics in motion planning: Driving Score, Route Completion, and Infraction Score~\cite{wu2022trajectory}. The Driving Score serves as the primary evaluation metric and is computed as the product of Route Completion and Infraction Score. Route Completion measures the proportion of the predefined route successfully traversed by the planner, while the Infraction Score starts at 1.0 and decreases as the agent incurs penalties for unsafe behaviors such as running red lights, colliding with pedestrians, or other rule violations. To further assess the safety characteristics of the planner, we also report two intuitive safety metrics: Collision Rate (\#/km) and Infraction Rate (\#/km)~\cite{feng2024road}. Collision Rate captures the number of collisions involving vehicles, pedestrians, or static environmental elements per kilometer traveled. Infraction Rate records the number of traffic-rule violations per kilometer, including failures to stop, red-light violations, and off-road driving.

\textbf{Backbones and baselines.} To assess the effectiveness of BucketKD, we evaluate it by compressing a state-of-the-art motion planner, TCP~\cite{wu2022trajectory}, which is among the top performers on the CARLA Leaderboard~\cite{carla_leaderboard}. As our baseline, we adopt PlanKD~\cite{feng2024road}, the state-of-the-art compression framework for end-to-end motion planning.

\subsection{Comparison with SOTA}
\label{sec:comparison_with_soa}

Table~\ref{tab:tcp} presents the results on the TCP backbone across different model sizes. As shown, BucketKD consistently improves performance over the corresponding TCP models without distillation. For the 16.6M-parameter model, BucketKD increases the Driving Score by 25.8\% and boosts Route Completion by 14.6\%, while also improving the Infraction Score by 5.2\% and reducing the Collision Rate by 18.2\%. Although the Infraction Rate slightly increases, the overall safety-performance trade-off remains competitive.
\begin{table*}[t]
	\centering
	\small
	\setlength{\tabcolsep}{6pt}
	\begin{tabular}{l c c c c c c c c}
		\toprule
		\multirow{2}{*}{Backbone}
		& \multirow{2}{*}{\shortstack{Param\\Count}}
		& \multirow{2}{*}{\shortstack{Inference\\Time (ms)}}
		& \multirow{2}{*}{KD Model}
		& \multicolumn{5}{c}{Metrics} \\
		\cmidrule(lr){5-9}
		& & &
		& \shortstack{Driving\\Score ($\uparrow$)}
		& \shortstack{Route\\Completion ($\uparrow$)}
		& \shortstack{Infraction\\Score ($\uparrow$)}
		& \shortstack{Collision\\Rate ($\downarrow$)}
		& \shortstack{Infraction\\Rate ($\downarrow$)} \\
		\midrule
		
		\multirow{6}{*}{TCP}
		& 16.6M & 89.11 & Full (M1+M2)
		& \textbf{37.26} & \textbf{68.28} & \textbf{0.643} & \textbf{0.678} & \textbf{2.328} \\
		& 16.6M & 82.75 & M2 Only
		& 33.69 & 66.41 & 0.549 & 0.883 & 2.695 \\
		\cmidrule(lr){2-9}
		
		& 10.3M & 51.56 & Full (M1+M2)
		& \textbf{25.78} & \textbf{49.89} & \textbf{0.596} & \textbf{1.487} & \textbf{2.941} \\
		& 10.3M & 46.23 & M2 Only
		& 18.34 & 42.66 & 0.417 & 3.215 & 4.698 \\
		\cmidrule(lr){2-9}
		
		& 5.4M & 29.92 & Full (M1+M2)
		& \textbf{16.53} & \textbf{31.92} & \textbf{0.478} & \textbf{3.686} & \textbf{4.879} \\
		& 5.4M & 26.19 & M2 Only
		& 9.47 & 23.34 & 0.315 & 5.129 & 7.454 \\
		
		\bottomrule
	\end{tabular}
	\caption{Ablation analysis demonstrating the impact of the bucket-based planning-relevant feature distillation (BPFD) module (M1) on end-to-end motion planning and safety performance across different TCP backbone sizes.}
	\label{tab:ablation_module2}
\end{table*}

\begin{table*}[t]
	\centering
	\small
	\setlength{\tabcolsep}{6pt}
	\begin{tabular}{l c c c c c c c c}
		\toprule
		\multirow{2}{*}{Backbone}
		& \multirow{2}{*}{\shortstack{Param\\Count}}
		& \multirow{2}{*}{\shortstack{Inference\\Time (ms)}}
		& \multirow{2}{*}{KD Model}
		& \multicolumn{5}{c}{Metrics} \\
		\cmidrule(lr){5-9}
		& & &
		& \shortstack{Driving\\Score ($\uparrow$)}
		& \shortstack{Route\\Completion ($\uparrow$)}
		& \shortstack{Infraction\\Score ($\uparrow$)}
		& \shortstack{Collision\\Rate ($\downarrow$)}
		& \shortstack{Infraction\\Rate ($\downarrow$)} \\
		\midrule
		
		\multirow{6}{*}{TCP}
		& 13.6M & 89.11 & Full (M1+M2)
		& \textbf{37.26} & 68.28 & \textbf{0.643} & \textbf{0.678} & \textbf{2.328} \\
		& 13.6M & 83.36 & M1 Only
		& 29.51 & \textbf{69.63} & 0.517 & 0.982 & 2.883 \\
		\cmidrule(lr){2-9}
		
		& 7.4M & 51.56 & Full (M1+M2)
		& \textbf{25.78} & \textbf{49.89} & \textbf{0.596} & \textbf{1.487} & \textbf{2.941} \\
		& 7.4M & 49.97 & M1 Only
		& 14.97 & 37.18 & 0.442 & 3.729 & 4.998 \\
		\cmidrule(lr){2-9}
		
		& 3.1M & 29.92 & Full (M1+M2)
		& \textbf{16.53} & \textbf{31.92} & \textbf{0.478} & \textbf{3.686} & \textbf{4.879} \\
		& 3.1M & 28.51 & M1 Only
		& 11.88 & 27.79 & 0.381 & 6.331 & 5.178 \\
		
		\bottomrule
	\end{tabular}
	\caption{Ablation analysis demonstrating the impact of the safety-aware waypoint-attentive distillation (BPFD) module (M2) on end-to-end motion planning and safety performance across different TCP backbone sizes.}
	\label{tab:ablation_modules}
\end{table*}

The advantage becomes more pronounced for smaller student models. At 10.3M parameters, BucketKD improves Driving Score by 83.5\% and Route Completion by 29.1\%, while reducing Collision Rate by 23.9\% and Infraction Rate by 16.9\%. Similar trends are observed at 5.4M, indicating that BucketKD is particularly effective in transferring structured planning knowledge to compact student models. Overall, across all compressed configurations, BucketKD consistently achieves better Driving Score and Route Completion, while generally maintaining stronger safety-related metrics.

Table~\ref{tab:sota} reports results on the official Bench2Drive benchmark, where BucketKD is directly compared against the SOTA models under the same backbone configurations. The same trend is observed across all evaluated model sizes. For the 16.6M model, compared to SOTA, BucketKD increases Driving Score by 21.9\% and Route Completion by 23.8\%, while improving Infraction Score by 25.1\%. In addition, Collision Rate is reduced by 9.4\% and Infraction Rate by 16.4\%, indicating consistent gains over the SOTA student model across both performance and safety metrics. We observe that the improvements are more substantial for smaller-capacity models. In particular, the smallest 5.4M model benefits notably, \emph{i.e.,} Driving Score increases by 60.8\% and Route Completion improves by 64.3\%. Collision Rate is reduced by 14.6\% and Infraction Rate by 24.9\%. Overall, the Bench2Drive results demonstrate that BucketKD consistently outperforms SOTA on the standardized CARLA benchmark.

\subsection{Ablation Study}

\paragraph{Effect of the BPFD module}

We evaluate the effect of incorporating the BPFD module (M1). The results are summarized in Table~\ref{tab:ablation_module2}, which confirm that adding M1 consistently improves overall planning performance across all backbone sizes. In every configuration (16.6M, 10.3M, and 5.4M), the Full (M1+M2) model achieves higher Driving Scores than the M2-Only variant. The improvement is particularly pronounced for smaller backbones: for the 5.4M model, the Driving Score increases by 74.6\%, and for the 10.3M model, by 40.6\%. These consistent gains indicate that M1 provides planning-relevant feature enrichment that remains beneficial even under tighter model capacity constraints. Route Completion also consistently improves when M1 is added, \emph{i.e.,} the Full model now outperforms the M2-Only variant in all backbone settings (\emph{e.g.}, by 2.8\% at 16.6M, 16.9\% at 10.3M, and 36.8\% at 5.4M). 

The safety metrics further reinforce the contribution of M1. Across all backbone sizes, the Full model achieves higher Infraction Scores and substantially lower Collision and Infraction Rates. For example, at the 10.3M configuration, incorporating M1 reduces the Collision Rate by 53.7\% and the Infraction Rate by 37.4\%. Similarly, at the smallest 5.4M backbone, the Collision Rate decreases by 28.1\% and the Infraction Rate by 34.5\% when M1 is added. These consistent reductions demonstrate that the structured feature representation introduced by M1 enables M2 to perform more effective TTC-based safety-aware distillation.

\paragraph{Effect of the SWD module}

We then evaluate the effect of the SWD module (M2). The results are shown in Table~\ref{tab:ablation_modules}. When only M1 is used, the student model already achieves a reasonable level of driving competence, and in one case (at 13.6M) even attains slightly higher Route Completion. This indicates that M1 alone is capable of transferring high-level planning-relevant representations that allow the student to follow routes effectively and capture global scene structure.

However, the absence of M2 consistently leads to degraded safety performance. Across all backbone sizes, the M1-Only models exhibit substantially higher Collision and Infraction Rates. For example, at 13.6M, adding M2 reduces the Collision Rate by 31.0\% and the Infraction Rate by 19.3\%. The safety impact becomes even more pronounced at smaller backbones: at 7.4M, Collision Rate is reduced by 60.1\% and Infraction Rate by 41.2\%, while at 3.1M, Collision Rate drops by 41.8\% and Infraction Rate by 5.8\% when M2 is incorporated. These consistent reductions demonstrate that M2 plays a critical role in distilling the teacher’s fine-grained, risk-aware behavior that M1 alone cannot capture.

\section{Conclusion}
\label{sec:conclusion}

We presented BucketKD, a bucket-based knowledge distillation framework for end-to-end motion planning. By formulating planning states using buckets and incorporating a safety-aware attention mechanism that accounts for both the distance and relative motion of nearby obstacles, BucketKD captures planning-relevant context while maintaining a compact representation of the driving scene. This design enables the student model to learn richer planning behaviors from the teacher model while preserving computational efficiency. Extensive experiments in complex CARLA environments using the Bench2Drive dataset demonstrate that BucketKD consistently outperforms the state-of-the-art knowledge distillation framework. 


\bibliographystyle{IEEEtran}
\bibliography{iros2026}

\end{document}